\def\year{2022}\relax
\documentclass[letterpaper]{article} 
\usepackage{aaai22}  
\usepackage{times}  
\usepackage{helvet}  
\usepackage{courier}  
\usepackage[hyphens]{url}  
\usepackage{graphicx} 
\usepackage{natbib}  
\usepackage{caption} 
\DeclareCaptionStyle{ruled}{labelfont=normalfont,labelsep=colon,strut=off} 
\usepackage{algorithm}
\usepackage{algorithmic}

\usepackage{newfloat}
\usepackage{listings}
\floatstyle{ruled}
\newfloat{listing}{tb}{lst}{}
\floatname{listing}{Listing}
\usepackage{booktabs} 
\usepackage{paralist}
\usepackage{amsthm}

\usepackage[shortlabels, inline]{enumitem}
\usepackage[np, autolanguage]{numprint}
\usepackage{multirow}
\usepackage[official]{eurosym}
\usepackage{xargs}   
\usepackage{xcolor}
\usepackage{amssymb}
\usepackage{amsmath}
\usepackage{relsize}
\usepackage{bbm}
\usepackage{subcaption}
\usepackage{acronym}
\acrodef{ML}{Machine Learning}
\acrodef{OCE}{Optimizable Counterfactual Explanations}
\acrodef{DATE}{Differentiable Approximations of Tree Ensembles}
\acrodef{GDPR}{General Data Protection Regulation}
\acrodef{FT}{Feature Tweaking}
\acrodef{RP}{Random Perturbation}
\usepackage[skip=0pt]{caption}
\newcommand{\OurMethod}{FOCUS}

\newcommand{\dubbelop}{$^{\blacktriangle}$}

\newcommand{\dubbelneer}{$^{\blacktriangledown}$}
\newcommand{\notsig}{$^{\circ}$}

\newcommand{\NoExample}{$^\otimes$}

\title{FOCUS: Flexible Optimizable Counterfactual Explanations for Tree Ensembles}
\author {
    Ana Lucic,\textsuperscript{\rm 1}
    Harrie Oosterhuis,\textsuperscript{\rm 2}
    Hinda Haned,\textsuperscript{\rm 1}
    Maarten de Rijke\textsuperscript{\rm 1}
}
\affiliations {
    \textsuperscript{\rm 1} University of Amsterdam\qquad
    \textsuperscript{\rm 2} Radboud University\\
    a.lucic@uva.nl, harrie.oosterhuis@ru.nl, h.haned@uva.nl, m.derijke@uva.nl
}

\begin{document}

\maketitle

\begin{abstract}
Model interpretability has become an important problem in \ac{ML} due to the increased effect that algorithmic decisions have on humans. 
Counterfactual explanations can help users understand not only why ML models make certain decisions, but also how these decisions can be changed. 
We frame the problem of finding counterfactual explanations as a gradient-based optimization task and extend previous work that could only be applied to differentiable models. 
In order to accommodate non-differentiable models such as tree ensembles, we use probabilistic model approximations in the optimization framework.
We introduce an approximation technique that is effective for finding counterfactual explanations for predictions of the original model and show that our counterfactual examples are significantly closer to the original instances than those produced by other methods specifically designed for tree ensembles. 
\end{abstract}

\section{Introduction}
\label{section:intro}
As \ac{ML} models are prominently applied and their outcomes have a substantial effect on the general population, there is an increased demand for understanding what contributes to their predictions \citep{doshi-2017-towards}. 
For an individual who is affected by the predictions of these models, it would be useful to have an \emph{actionable} explanation -- one that provides insight into how these decisions can be \emph{changed}. 
The General Data Protection Regulation (GDPR) is an example of recently enforced regulation in Europe which gives an individual the right to an explanation for algorithmic decisions, making the interpretability problem a crucial one for organizations that wish to adopt more data-driven decision-making processes \citep{gdpr}. 

Counterfactual explanations are a natural solution to this problem since they frame the explanation in terms of what input (feature) changes are required to change the output (prediction). 
For instance, a user may be denied a loan based on the prediction of an \ac{ML} model used by their bank. 
A counterfactual explanation could be: ``\textit{Had your income been \euro  $1000$ higher, you would have been approved for the loan}.''
We focus on finding \emph{optimal} counterfactual explanations: the \emph{minimal} changes to the input required to change the outcome. 

Counterfactual explanations are based on counterfactual examples: generated instances that are close to an existing instance but have an alternative prediction. 
The difference between the original instance and the counterfactual example is the counterfactual explanation. 
\citet{wachter_counterfactual_2017} propose framing the problem as an optimization task, but their work assumes that the underlying machine learning models are differentiable, which excludes an important class of widely applied and highly effective non-differentiable models: tree ensembles. 
We propose a method that relaxes this assumption and builds upon the work of \citeauthor{wachter_counterfactual_2017} by introducing differentiable approximations of tree ensembles that can be used in such an optimization framework. 
Alternative non-optimization approaches for generating counterfactual explanations for tree ensembles involve an extensive search over many possible paths in the ensemble that could lead to an alternative prediction \citep{tolomei_interpretable_2017}. 

Given a trained tree-based model $f$, we probabilistically approximate $f$ by replacing each split in each tree with a sigmoid function centred at the splitting threshold. If $f$ is an ensemble of trees, then we also replace the maximum operator with a softmax. 
This approximation allows us to generate a counterfactual example $\bar{x}$ for an instance $x$ based on the minimal perturbation of $x$ such that the prediction changes: $y_{x} \neq y_{\bar{x}}$, where $y_{x}$ and $y_{\bar{x}}$ are the labels $f$ assigns to $x$ and $\bar{x}$, respectively. 
This leads us to our main research question:
\begin{quote}
\emph{Are counterfactual examples generated by our method closer to the original input instances than those generated by existing heuristic methods?}
\end{quote}
Our main findings are that our method is
\begin{enumerate*}[label=(\roman*)]
\item a more \emph{effective} counterfactual explanation method for tree ensembles than previous approaches since it manages to produce counterfactual examples that are closer to the original input instances than existing approaches; 
\item a more \emph{efficient} counterfactual explanation method for tree ensembles since it is able to handle larger models than existing approaches; and
\item a more \emph{reliable} counterfactual explanation method for tree ensembles since it is able to generate counterfactual explanations for all instances in a dataset, unlike existing approaches specific to tree ensembles. 
\end{enumerate*}

\section{Related Work}
\label{section:relatedwork}
\subsection{Counterfactual Explanations}
Counterfactual examples have been used in a variety of ML areas, such as reinforcement learning \citep{madumal_explainable_2019}, deep learning \citep{alaa_deep_2017}, and explainable AI (XAI). 
Previous XAI methods for generating counterfactual examples are either model-agnostic \citep{poyiadzi_face_2020, karimi_model-agnostic_2019, laugel_inverse_2017, van_looveren_interpretable_2020,  mothilal_explaining_2020} or model-specific \citep{wachter_counterfactual_2017, grath_interpretable_2018, tolomei_interpretable_2017, kanamori_dace_2020, russell_efficient_2019, dhurandhar_explanations_2018}. 
Model-agnostic approaches treat the original model as a ``black-box'' and only assume query access to the model, whereas model-specific approaches typically do not make this assumption and can therefore make use of its inner workings. 
Our work is a model-specific approach for generating counterfactual examples through optimization. 
Previous model-specific work for generating counterfactual examples through optimization has solely been conducted on differentiable models \citep{wachter_counterfactual_2017, grath_interpretable_2018, dhurandhar_explanations_2018}. 

\subsection{Algorithmic Recourse}
Algorithmic recourse is a line of research that is closely related to counterfactual explanations, except that these methods include the additional restriction that the resulting explanation must be \emph{actionable} \citep{ustun_actionable_2019, joshi_towards_2019, karimi_recourse_2020, karimi_imperfect_causal_2020}. 
This is done by selecting a subset of the features to which perturbations can be applied in order to avoid explanations that suggest impossible or unrealistic changes to the feature values (i.e., change \textit{age} from \numprint{50} $\to$ \numprint{25} or change \textit{marital\_status} from $\mathrm{MARRIED}$ $\to$ UNMARRIED). 
Although this work has produced impressive theoretical results, it is unclear how realistic they are in practice, especially for complex ML models such as tree ensembles. 
Existing algorithmic recourse methods cannot solve our task because they 
\begin{enumerate*}[label=(\roman*)]
	\item are either restricted to solely linear \citep{ustun_actionable_2019} or  differentiable \citep{joshi_towards_2019} models, or
	\item  require access to causal information \citep{karimi_recourse_2020, karimi_imperfect_causal_2020}, which is rarely available in real world settings. 
\end{enumerate*}

\subsection{Adversarial Examples}
Adversarial examples are a type of counterfactual example with the additional constraint that the minimal perturbation results in an alternative prediction that is \emph{incorrect}. 
There are a variety of methods for generating adversarial examples \cite{goodfellow_explaining_2015,szegedy_intriguing_2014,su_one_2019,brown_adversarial_2018}; a more complete overview can be found in \cite{biggio_wild_2018}. 
The main difference between adversarial examples and counterfactual examples is in the intent: adversarial examples are meant to \emph{fool} the model, whereas counterfactual examples are meant to \emph{explain} the model.

\subsection{Differentiable Tree-based Models}
Part of our contribution involves constructing differentiable versions of tree ensembles by replacing each splitting threshold with a sigmoid function. 
This can be seen as using a (small) neural network to obtain a smooth approximation of each tree. 
Neural decision trees \citep{balestriero_neural_2017, yang_deep_2018} are also differentiable versions of trees, which use a full neural network instead of a simple sigmoid. 
However, these do not optimize for approximating an already trained model. Therefore, unlike our method, they are not an obvious choice for finding counterfactual examples for an existing model. 
Soft decision trees~\citep{hinton_distilling_2014} are another example of differentiable trees, which instead approximate a neural network with a decision tree. 
This can be seen as the inverse of our task.

\section{Problem Definition}
\label{section:problem-definition}

A \emph{counterfactual explanation} for an instance $x$ and a model $f$, $\Delta_{x}$, is a minimal perturbation of $x$ that changes the prediction of $f$. 
$f$ is a probabilistic classifier, where $f(y\mid x)$ is the probability of $x$ belonging to class $y$ according to $f$.
The prediction of $f$ for $x$ is the most probable class label $y_x = \arg\max_{y} f(y \mid x)$, and
a perturbation $\bar{x}$ is a counterfactual example for $x$ if, and only if, $y_x \not = y_{\bar{x}}$, that is:
\begin{align}
\arg\max_{y} f(y \mid x)
\not =
\arg\max_{y'} f(y' \mid \bar{x}).
\label{eq:cfexample}
\end{align}
In addition to changing the prediction, the distance between $x$ and $\bar{x}$ should also be minimized. 
We therefore define an \emph{optimal counterfactual example} $\bar{x}^*$ as: 
\begin{equation}
 \bar{x}^* := \arg\min_{\bar{x}} d(x, \bar{x}) 
 \text{ such that }
y_x \not = y_{\bar{x}}.
\label{eq:optimalcondition}
\end{equation}
\noindent
where $d(x, \bar{x})$ is a differentiable distance function. 
The corresponding \emph{optimal counterfactual explanation} $\Delta^*_{x}$ is:
\begin{align}
\Delta^*_{x} = \bar{x}^* - x.
\end{align} 

This definition aligns with previous \ac{ML} work on counterfactual explanations \citep{laugel_inverse_2017, karimi_model-agnostic_2019, tolomei_interpretable_2017}. 
We note that this notion of \emph{optimality} is purely from an algorithmic perspective and does not necessarily translate to optimal changes in the real world, since the latter are completely dependent on the context in which they are applied. 
It should be noted that if the loss space is non-convex, it is possible that more than one optimal counterfactual explanation exists.

Minimizing the distance between $x$ and $\bar{x}$ should ensure that $\bar{x}$ is as close to the decision boundary as possible. 
This distance indicates the effort it takes to apply the perturbation in practice, and an optimal counterfactual explanation shows how a prediction can be changed with the least amount of effort.
An optimal explanation provides the user with interpretable and potentially actionable feedback related to understanding the predictions of model $f$.

\citet{wachter_counterfactual_2017} recognized that counterfactual examples can be found through gradient descent if the task is cast as an optimization problem.
Specifically, they use a loss consisting of two components: 
\begin{enumerate*}[label=(\roman*)]
	\item a prediction loss to change the prediction of $f$: $\mathcal{L}_{pred}(x, \bar{x} \mid f)$, and
	\item a distance loss to minimize the distance $d$: $\mathcal{L}_{dist}(x, \bar{x} \mid d)$.
\end{enumerate*}
The complete loss is a linear combination of these two parts, with a weight $\beta \in \mathbb{R}_{>0}$:
\begin{align}
\label{eq:mainloss}
\mathcal{L}(x, \bar{x} \mid f, d) = \mathcal{L}_{pred}(x, \bar{x} \mid f) + \beta \mathcal{L}_{dist}(x, \bar{x} \mid d).
\end{align}
The assumption here is that an optimal counterfactual example $\bar{x}^*$ can be found by minimizing the overall loss:
\begin{align}
\bar{x}^* = \arg\min_{\bar{x}} \mathcal{L}(x, \bar{x} \mid f, d).
\end{align}
\citet{wachter_counterfactual_2017} propose a prediction loss $\mathcal{L}_{pred}$ based on the mean-squared-error. 
A clear limitation of this approach is that it assumes $f$ is differentiable.
This excludes many commonly used \ac{ML} models, including tree-based models, on which we focus in this paper.

\section{Method}
\label{section:method}
To mimic many real-world scenarios, we assume there exists a trained model $f$ that we need to explain. The goal here is not to create a new, inherently interpretable tree-based model, but rather to explain a model that already exists.

\subsection{Loss Function Definitions}

We use a hinge-loss since we assume a classification task:
\begin{align}
\begin{split}
& \mbox{}\hspace*{-1mm}\mathcal{L}_{pred}(x, \bar{x} \mid f) = \\
& \mbox{}\hspace*{-1mm}{\mathbbm{1}}\left[\arg\max_{y} f(y \mid x) = \arg\max_{y'} f(y' \mid \bar{x})\right] \cdot  f(y' \mid \bar{x}).
\end{split}
\end{align}
Allowing for flexibility in the choice of distance function allows us to tailor the explanations to the end-users' needs. We make the preferred notion of \emph{minimality} explicit through the choice of distance function. 
Given a differentiable distance function $d$, the distance loss is: 
\begin{align}
\mathcal{L}_{dist}(x, \bar{x}) = d(x, \bar{x}). 
\end{align}
Building off of \citet{wachter_counterfactual_2017}, we propose incorporating differentiable approximations of non-differentiable models to use in the gradient-based optimization framework. 
Since the approximation $\tilde{f}$ is derived from the original model $f$, it should match $f$ closely: $\tilde{f}(y \mid x) \approx f(y \mid x)$. 
We define the approximate prediction loss as follows:
\begin{align}
\begin{split}
&\mbox{}\hspace*{-2mm}\widetilde{\mathcal{L}}_{pred}(x, \bar{x} \mid f, \tilde{f}) = \\
&\mbox{}\hspace*{-2mm} \mathbbm{1}\left[\arg\max_{y} f(y \mid x) = \arg\max_{y'} f(y' \mid \bar{x})\right] \cdot  \tilde{f}(y' \mid \bar{x}).
\end{split}
\end{align}
This loss is based both on the original model $f$ and the approximation $\tilde{f}$:
the loss is active as long as the prediction according to $f$ has not changed, but its gradient is based on the differentiable $\tilde{f}$. 
This prediction loss encourages the perturbation to have a different prediction than the original instance by penalizing an unchanged instance. 
The approximation of the complete loss becomes:
\begin{equation}
\widetilde{\mathcal{L}}(x, \bar{x} \mid f, \tilde{f}, d) =\widetilde{\mathcal{L}}_{pred}(x, \bar{x} \mid f, \tilde{f}) + \beta \cdot \mathcal{L}_{dist}(x, \bar{x} \mid d).
\label{eq:approxloss}
\end{equation}
Since we assume that it approximates the complete loss, 
\begin{align}
\widetilde{\mathcal{L}}(x, \bar{x} \mid f, \tilde{f}, d) \approx \mathcal{L}(x, \bar{x} \mid f, d),
\end{align}
we also assume that an optimal counterfactual example can be found by minimizing it:
\begin{align}
\bar{x}^* \approx \arg\min_{\bar{x}} \, \widetilde{\mathcal{L}}(x, \bar{x} \mid f, \tilde{f}, d).
\label{eq:xbar}
\end{align}

\begin{figure}[t]
\centering
\includegraphics[scale=0.6]{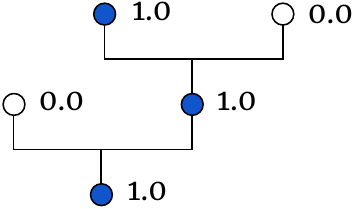} 
\quad
\includegraphics[scale=0.6]{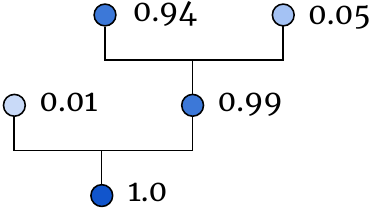}
\caption{
Left: A decision tree $\mathcal{T}$ and node activations for a single instance. Right: a differentiable approximation of the same tree $\widetilde{\mathcal{T}}$ and activations for the same instance.
}
\label{fig:exampletrees}
\end{figure}

\subsection{Tree-based Models}
To obtain the differentiable approximation $\tilde{f}$ of $f$, we construct a probabilistic approximation of the original tree ensemble $f$.
Tree ensembles are based on decision trees; a single decision tree $\mathcal{T}$ uses a binary-tree structure to make predictions about an instance $x$ based on its features.
Figure~\ref{fig:exampletrees} shows a simple decision tree consisting of five nodes.
A node $j$ is activated if its parent node $p_j$ is activated and feature $x_{f_j}$ is on the correct side of the threshold $\theta_j$; which side is the correct side depends on whether $j$ is a \emph{left} or \emph{right} child; with the exception of the root node which is always activated.
Let $t_j(x)$ indicate if node $j$ is activated:
\begin{equation}
\mbox{}\hspace*{-2mm}t_j(x) =
    \begin{cases}
   1, & \text{if $j$ is the root}, \\
   t_{p_j}(x) \cdot  \mathbbm{1}[x_{f_j} > \theta_j], & \text{if $j$ is a left child}, \\
    t_{p_j}(x) \cdot  \mathbbm{1}[x_{f_j} \leq \theta_j], &\text{if $j$ is a right child}.
    \end{cases}
\end{equation}
$\forall x, \,  t_0(x) = 1$.
Nodes that have no children are called \emph{leaf nodes}; an instance $x$ always ends up in a single leaf node.
Every leaf node $j$ has its own predicted distribution $\mathcal{T}(y \mid j)$; the prediction of the full tree is given by its activated leaf node. 
Let $\mathcal{T}_{\textit{leaf}}$ be the set of leaf nodes in $\mathcal{T}$, then:
\begin{equation}
(j \in \mathcal{T}_{\textit{leaf}} \land t_j(x) = 1) \rightarrow \mathcal{T}(y \mid x) = \mathcal{T}(y \mid j).
\end{equation}
Alternatively, we can reformulate this as a sum over leaves:
\begin{equation}
\mathcal{T}(y \mid x) = \sum_{j \in \mathcal{T}_\mathit{leaf}}  t_j(x) \cdot \mathcal{T}(y \mid j).
\end{equation}
Generally, tree ensembles are deterministic; let $f$ be an ensemble of $M$ many trees with weights $\omega_m \in \mathbb{R}$, then:
\begin{equation}
f(y \mid x)
=  \arg\max_{y'} \sum_{m=1}^M \omega_m \cdot \mathcal{T}_m(y' \mid x).
\end{equation}

\begin{figure*}[h]
\centering
\includegraphics[scale=0.3]{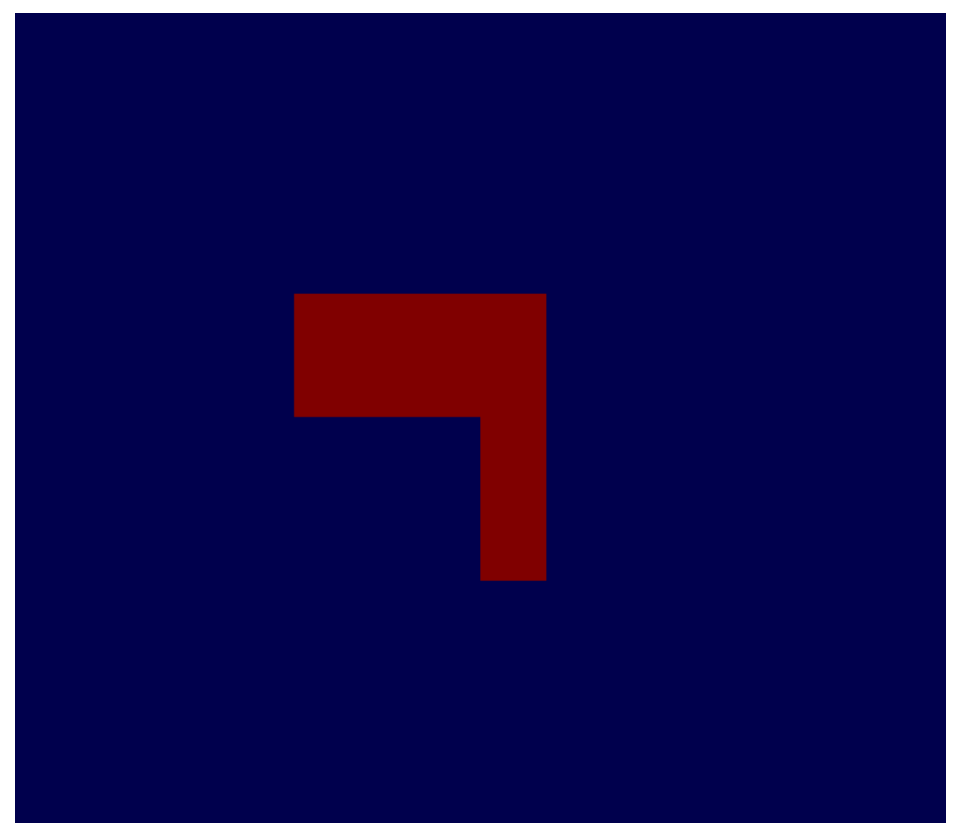} 
\includegraphics[scale=0.3]{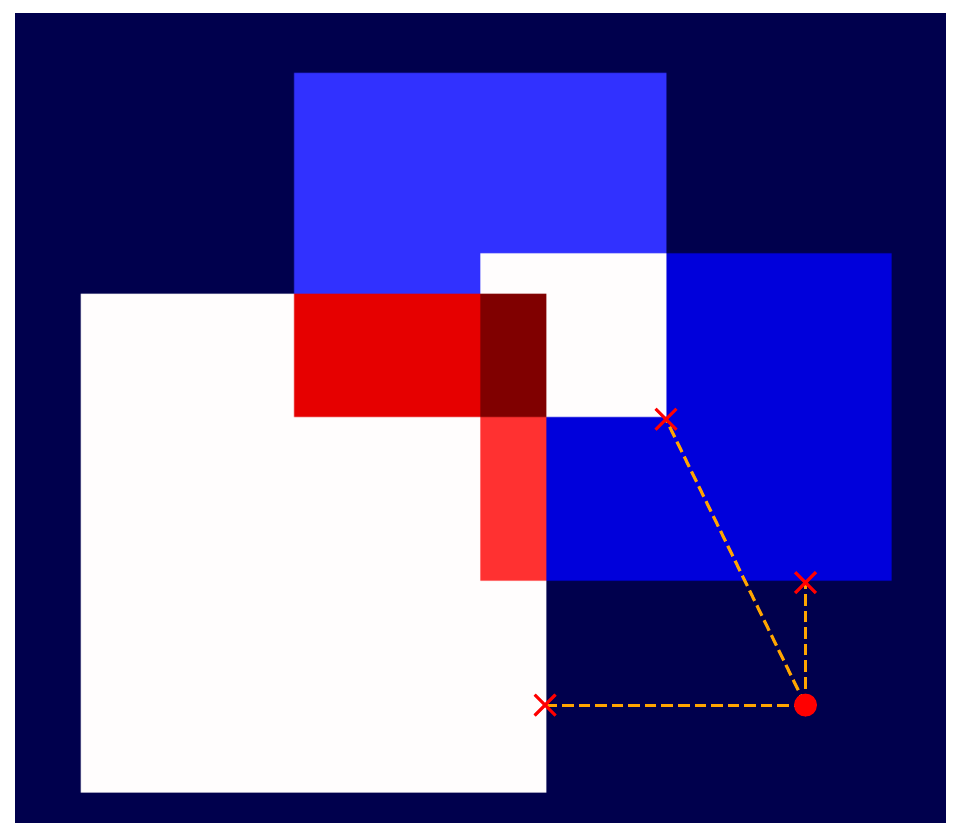} 
\includegraphics[scale=0.3]{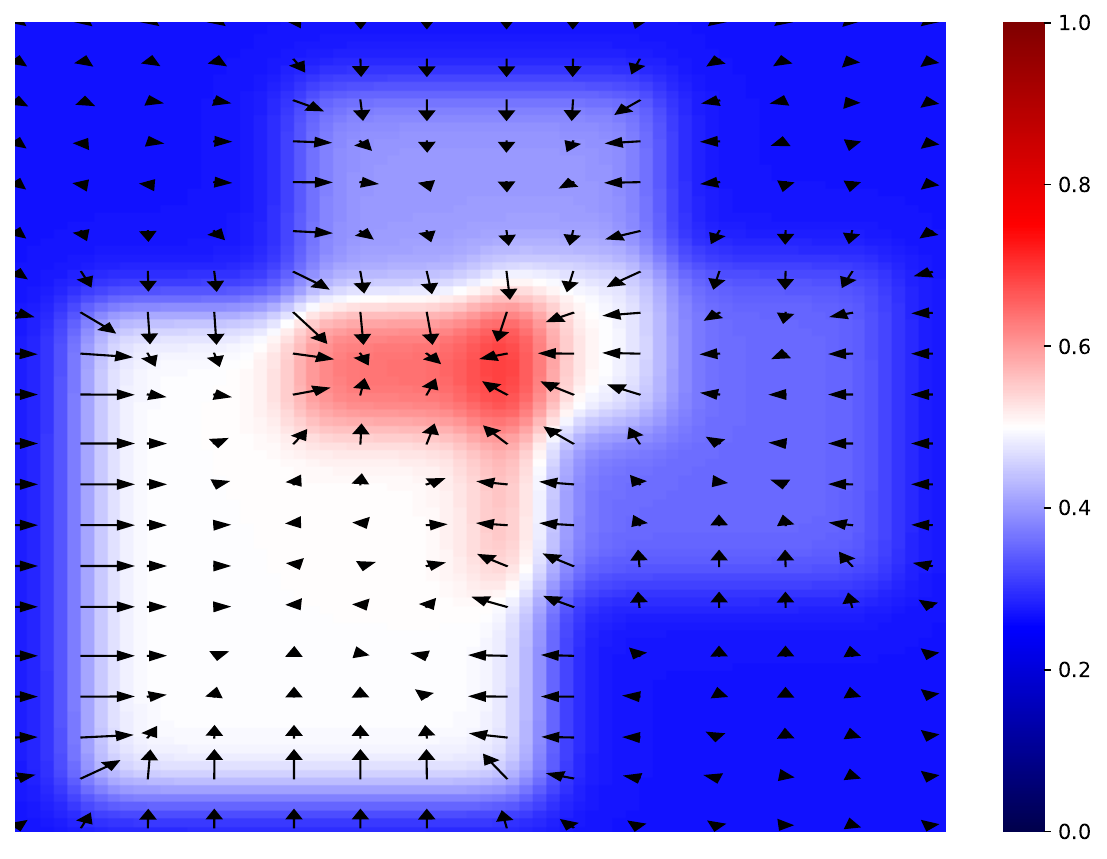}
\caption{
An example of how the \ac{FT} baseline method (explained in Section~\ref{section:baselineft}) and our method handle an adaptive boosting ensemble with three trees.
Left: decision boundary of the ensemble.
Middle: three positive leaves that form the decision boundary, an example instance and the perturbed examples suggested by \acs{FT}. 
Right: approximated loss $\widetilde{\mathcal{L}}_{pred}$ and its gradient w.r.t. $\bar{x}$. 
The \acs{FT} perturbed examples do not change the prediction of the forest, whereas the gradient of the differentiable approximation leads toward the true decision boundary.
}
\label{fig:approxensemble}
\end{figure*}

\subsection{Approximations of Tree-based Models}
If $f$ is not differentiable, we are unable to calculate its gradient with respect to the input $x$. 
However, the non-differentiable operations in our formulation are 
\begin{enumerate*}[label=(\roman*)]
	\item the indicator function, and
	\item a maximum operation, 
\end{enumerate*}
both of which can be approximated by differentiable functions.
First, we introduce the $\widetilde{t}_j(x)$ function that \emph{approximates the activation of node} $j$: $\widetilde{t}_j(x) \approx t_j(x)$, using a sigmoid function with parameter $\sigma \in \mathbb{R}_{>0}$:
$
\textit{sig}(z) = (1 + \exp(\sigma \cdot z))^{-1}
$
and
\begin{align}
\widetilde{t}_j(x) &{} =
    \begin{cases}
    1, & \text{if $j$ is the root}, \\
   \widetilde{t}_{p_j}(x) \cdot \textit{sig}(\theta_j {-} x_{f_j}), & \text{if $j$ is left child}, \\
   \widetilde{t}_{p_j}(x) \cdot  \textit{sig}( x_{f_j} {-} \theta_j), & \text{if $j$ is right child}.
    \end{cases}
\label{eq:sigma}    
\end{align}
As $\sigma$ increases, $\widetilde{t}_j$ approximates $t_j$ more closely.
Next, we introduce a \emph{tree approximation}:
\begin{equation}
\widetilde{\mathcal{T}}(y \mid x) = \sum_{j \in \mathcal{T}_\mathit{leaf}}  \widetilde{t}_j(x) \cdot \mathcal{T}(y \mid j).
\end{equation}
The approximation $\widetilde{\mathcal{T}}$ uses the same tree structure and thresholds as $\mathcal{T}$.
However, its activations are no longer deterministic but instead are dependent on the distance between the feature values $x_{f_j}$ and the thresholds $\theta_j$.
Lastly, we replace the maximum operation of $f$ by a softmax with temperature $\tau\in\mathbb{R}_{>0}$, resulting in:
\begin{align}
\tilde{f}(y \mid x)
= \frac{
\exp\left(\tau \cdot \sum_{m=1}^M \omega_m \cdot \widetilde{\mathcal{T}}_m(y \mid x)\right)
}{
\sum_{y'} \exp\left(\tau \cdot \sum_{m=1}^M \omega_m \cdot \widetilde{\mathcal{T}}_m(y' \mid x)\right)
}.
\label{eq:tau}
\end{align}
The approximation $\tilde{f}$ is based on the original model $f$ and the parameters $\sigma$ and $\tau$.
This approximation is applicable to any tree-based model, and 
how well $\tilde{f}$ approximates $f$ depends on the choice of $\sigma$ and $\tau$.
The approximation is potentially perfect since
\begin{align}
\lim_{\sigma,\tau\rightarrow\infty}
\tilde{f}(y \mid x) = f(y \mid x).
\end{align}
\subsection{Our Method: FOCUS}
We call our method FOCUS: Flexible Optimizable CounterfactUal Explanations for Tree EnsembleS. 
It takes as input an instance $x$, a tree-based classifier $f$, and two hyperparameters: $\sigma$ and $\tau$ which we use to create the approximation $\tilde{f}$. 
Following Equation~\ref{eq:xbar}, FOCUS outputs the optimal counterfactual example $\bar{x}^*$, from which we derive the optimal counterfactual explanation $\Delta^*_{x} = \bar{x}^* - x$.

\subsection{Effects of Hyperparameters}
Increasing $\sigma$ in $\tilde{f}$ eventually leads to exact approximations of the indicator functions, while increasing $\tau$ in $\tilde{f}$ leads to a completely unimodal softmax distribution. 
It should be noted that our approximation $\tilde{f}$ is not intended to replace the original model $f$ but rather to create a differentiable version of $f$ from which we can generate counterfactual examples through optimization. 
In practice, the original model $f$ would still be used to make predictions and the approximation would solely be used to generate counterfactual examples.

\section{Experimental Setup}
\label{section:exp-setup}

We consider \numprint{42} experimental settings to find the best counterfactual explanations using FOCUS. 
We jointly tune the hyperparameters of FOCUS ($\sigma, \tau, \beta, \alpha$) using Adam~\citep{kingma_adam:_2017} for \numprint{1000} iterations. 
We choose the hyperparameters that produce
\begin{enumerate*}[label=(\roman*)]
	\item a valid counterfactual example for every instance in the dataset, and
	\item the smallest mean distance between corresponding pairs ($x$, $\bar{x}$).
\end{enumerate*}

We evaluate FOCUS on four binary classification datasets: \textit{Wine Quality} \citep{wine_2009}, \textit{HELOC} \citep{fico_2017}, \textit{COMPAS} \citep{compas-dataset-2017}, and \textit{Shopping} \citep{shoppingdataset}. 
For each dataset, we train three types of tree-based models: Decision Trees (DT), Random Forests (RF), and Adaptive Boosting Trees (AB) with DTs as the base learners. 
We compare against two baselines that generate counterfactual examples for tree ensembles based on the inner workings of the model: Feature Tweaking (FT) by \citet{tolomei_interpretable_2017} and Distribution-Aware Counterfactual Explanations (DACE) by \citet{kanamori_dace_2020}.

\subsection{Baseline: Feature Tweaking}
\label{section:baselineft}
Feature Tweaking identifies the leaf nodes where the prediction of the leaf nodes do not match the original prediction $y_x$: it recognizes the set of leaves that if activated, $t_j(\bar{x}) = 1$, would change the prediction of a tree $\mathcal{T}$:
\begin{equation}
\mathcal{T}_\textit{change} = \left\{ j \mid j \in   \mathcal{T}_\textit{leaf} \land y_x \not = \arg \max_y T(y\mid j) \right\}.
\end{equation}
For every $\mathcal{T}$ in $f$, \ac{FT} generates a perturbed example per node in $\mathcal{T}_\textit{change}$ so that it is activated with at least an $\epsilon$ difference per threshold, and then selects the most optimal example (i.e., the one closest to the original instance).
For every feature threshold $\theta_j$ involved, the corresponding feature is perturbed accordingly: $\bar{x}_{f_j} = \theta_j \pm \epsilon$.
The result is a perturbed example that was changed minimally to activate a leaf node in $\mathcal{T}_\textit{change}$. 
In our experiments, we test $\epsilon \in \{0.001, 0.005, 0.01, 0.1\}$, and choose the $\epsilon$ that minimizes the mean distance to the original input, while maximizing the number of counterfactual examples generated. 

The main problem with \ac{FT} is that the perturbed examples are not necessarily counterfactual examples, since changing the prediction of a single tree $\mathcal{T}$ does not guarantee a change in the prediction of the full ensemble $f$.
Figure~\ref{fig:approxensemble} shows all three perturbed examples generated by \ac{FT} for a single instance. 
In this case, none of the generated examples change the model prediction and therefore none are valid counterfactual examples. 

Figure~\ref{fig:approxensemble} shows how FOCUS and FT handle an adaptive boosting ensemble using a two-feature ensemble with three trees. 
On the left is the decision boundary for a standard tree ensemble; the middle visualizes the positive leaf nodes that form the decision boundary; on the right is the approximated loss $\widetilde{\mathcal{L}}_{pred}$ and its gradient w.r.t. $\bar{x}$.
The gradients push features close to thresholds harder and in the direction of the decision boundary if $\widetilde{\mathcal{L}}$ is convex.

\subsection{Baseline: DACE}
\label{section:baselinedace}
DACE generates counterfactual examples that account for the underlying data distribution through a novel cost function using Mahalanobis distance and a local outlier factor (LOF):
\begin{align}
\label{eq:daceloss}
\begin{split}
& d_\mathit{DACE}(x, \bar{x}|X, C) = \\
& {d_\mathit{Mahalanobis}}^2(x, \bar{x}|C) + \lambda q_k(x, \bar{x}|X), 
\end{split}
\end{align}
where $C$ is the covariance matrix, $q_k$ is the $k$-LOF \cite{breunig_lof_2020}, $X$ is the training set, and $\lambda$ is the trade-off parameter. 
The $k$-LOF measures the degree to which an instance is an outlier in the context of its $k$-nearest neighbors.\footnote{We use $k=1$ in our experiments, since this is the value of $k$ that is supported in the code kindly provided to us by the authors, for which we are very grateful.}
To generate counterfactual examples, DACE formulates the task as a mixed-integer linear optimization problem and uses the CPLEX Optimizer\footnote{\url{http://www.ibm.com/analytics/cplex-optimizer}} to solve it. 
We refer the reader to the original paper for a more detailed overview of this cost function. 
The $q_k$ term in the loss function penalizes counterfactual examples that are outliers, and therefore decreasing $\lambda$ results in a greater number of counterfactual examples. 
In our experiments, we test $\lambda \in \{0.001, 0.01, 0.1, 0.5, 1.0\}$, and choose the $\lambda$ that minimizes the mean distance to the original input, while maximizing the number of counterfactual examples generated. 
The main issue with DACE is that even for very small values of $\alpha$, it is unable to generate counterfactual examples for the majority of instances in the test sets (see Table~\ref{table:experiment2}). 
The other issue is that it is unable to run on some of our models because the problem size is too large when using the free Python API of CPLEX.

\subsection{Datasets}
\label{section:datasets}
We evaluate \OurMethod{} on four binary classification tasks using the following datasets: \textit{Wine Quality} \citep{wine_2009}, \textit{HELOC} \citep{fico_2017}, COMPAS \citep{compas-dataset-2017}, and \textit{Shopping} \citep{shoppingdataset}. 
The \textit{Wine Quality} dataset (\numprint{4898} instances, \numprint{11} features) is about predicting the quality of white wine on a 0--10 scale. 
We adapt this to a binary classification setting by labelling the wine as ``high quality'' if the quality is $\geq$ \numprint{7}.
The \textit{HELOC} set (\numprint{10459} instances, \numprint{23} features) is from the Explainable Machine Learning Challenge at NeurIPS 2017, where the task is to predict whether or not a customer will default on their loan. 
The \textit{COMPAS} dataset (\numprint{6172} instances, \numprint{6} features) is used for detecting bias in ML systems, where the task is predicting whether or not a criminal defendant will reoffend upon release. 
The \textit{Shopping} dataset (\numprint{12330} instances, \numprint{9} features) entails predicting whether or not an online website visit results in a purchase. 
We scale all features such that their values are in the range $\left[0, 1\right]$ and remove categorical features. 

\subsection{Models}
We train three types of tree-based models on \numprint{70}\% of each dataset: Decision Trees (DTs), Random Forests (RFs), and Adaptive Boosting (AB) with DTs as the base learners. 
We use the remaining \numprint{30}\% to find counterfactual examples for this test set. 
In total we have \numprint{12} models (\numprint{4} datasets $\times$ \numprint{3} tree-based models). See Appendix A for more details.

\subsection{Evaluation Metrics}
\label{section:evaluation}
We evaluate the counterfactual examples produced by \OurMethod{} based on how close they are to the original input using three metrics. 
Mean distance, $d_\mathit{mean}$, measures the distance from the original input, averaged over all examples. 
Mean relative distance, $d_\mathit{Rmean}$, measures pointwise ratios of distance to the original input. 
This helps us interpret individual improvements over the baselines; if $d_\mathit{Rmean} < 1$, \OurMethod{}'s counterfactual examples are on average closer to the original input compared to the baseline. 
We also evaluate the proportion of \OurMethod{}'s counterfactual examples that are closer to the original input compared to the baselines ($\mathit{\%_{closer}}$). 
We test the metrics in terms of four distance functions: Euclidean, Cosine, Manhattan and Mahalanobis.

\begin{table*}[]
\begin{tabular*}{\textwidth}{l @{\extracolsep{\fill}} llrrrrrrrrrr}
\toprule
  &                   &                  & \multicolumn{3}{c}{\textbf{Euclidean}}                                                              & \multicolumn{3}{c}{\textbf{Cosine}}                                                                 & \multicolumn{3}{c}{\textbf{Manhattan}}                                                              \\
  \cmidrule(r){4-6}\cmidrule(r){7-9} \cmidrule(r){10-12}
 \textbf{Dataset}                 &     \textbf{Metric}                & \textbf{Method}                 & \multicolumn{1}{c}{\textbf{DT}} & \multicolumn{1}{c}{\textbf{RF}} & \multicolumn{1}{c}{\textbf{AB}} & \multicolumn{1}{c}{\textbf{DT}} & \multicolumn{1}{c}{\textbf{RF}} & \multicolumn{1}{c}{\textbf{AB}} & \multicolumn{1}{c}{\textbf{DT}} & \multicolumn{1}{c}{\textbf{RF}} & \multicolumn{1}{c}{\textbf{AB}} \\
 \midrule
\textit{}        &    $d_{mean}$                & \textit{FT}               & 0.269 & \textbf{0.174}  & 0.267\rlap\NoExample  & 0.030 & 0.017  & 0.034\rlap\NoExample  & 0.269 & \textbf{0.223} & 0.382\rlap\NoExample  \\
\textit{Wine}    &                    & \textit{\OurMethod{}}              & \textbf{0.268}\rlap{\notsig} & 0.188\rlap{\dubbelop}   & \textbf{0.188}\rlap{\dubbelneer}  & \textbf{0.003}\rlap{\dubbelneer} & \textbf{0.008}\rlap{\dubbelneer}  & \textbf{0.014}\rlap{\dubbelneer}  & \textbf{0.268}\rlap{\notsig} & 0.312\rlap{\dubbelop} & \textbf{0.360}\rlap{\dubbelneer}  \\
\cmidrule{2-12}
\textit{Quality}        &       $d_{Rmean}$             & \textit{\OurMethod{}/FT}           & 0.990 & 1.256  & 0.649  & 0.066 & 0.821  & 0.312  & 0.990 & 1.977 & 0.924  \\
\textit{}        &      $\mathit{\%_{closer}}$              & \textit{\OurMethod{} \textless FT}                           & 100\% & 21.0\% & 87.5\% & 100\% & 80.8\% & 95.1\% & 100\% & 5.4\% & 58.6\%                           \\

\midrule

                 &      $d_{mean}$              & \textit{FT}               & \textbf{0.120}  & 0.210  & 0.185  & 0.003  & 0.008  & 0.007  & \textbf{0.135}  & \textbf{0.278}  & \textbf{0.198}  \\
  \textit{HELOC}               &                    & \textit{\OurMethod{}}              & 0.133\rlap{\dubbelop}  & \textbf{0.186}\rlap{\dubbelneer}  & \textbf{0.136}\rlap{\dubbelneer}  & \textbf{0.001}\rlap{\dubbelneer}  & \textbf{0.002}\rlap{\dubbelneer}  & \textbf{0.001}\rlap{\dubbelneer}  & 0.152\rlap{\dubbelop}  & 0.284\rlap{\notsig}  & 0.203\rlap{\notsig}  \\
\cmidrule{2-12}

                 &     $d_{Rmean}$               & \textit{\OurMethod{}/FT}           & 1.169  & 0.942  & 0.907  & 0.303  & 0.285  & 0.421  & 1.252  & 1.144  & 1.364  \\ 
                 &     $\mathit{\%_{closer}}$               & \textit{\OurMethod{} \textless FT} & 16.6\% & 57.9\% & 71.9\% & 91.6\% & 91.5\% & 92.9\% & 51.3\% & 43.6\% & 24.2\%          \\
 
 \midrule

                 &      	$d_{mean}$              & \textit{FT}               & \textbf{0.082} & \textbf{0.075} & 0.081 & 0.013 & 0.014 & 0.015 & \textbf{0.086} & \textbf{0.078} & \textbf{0.085} \\
  \textit{COMPAS}               &              & \textit{\OurMethod{}}              & 0.092\rlap{\dubbelop} & 0.079\rlap{\notsig} & \textbf{0.076}\rlap{\dubbelneer} & \textbf{0.008}\rlap{\dubbelneer} & \textbf{0.011}\rlap{\dubbelneer} & \textbf{0.007}\rlap{\dubbelneer} & 0.093\rlap{\dubbelop} & 0.085\rlap{\notsig} & 0.090\rlap{\notsig} \\
\cmidrule{2-12}
                 &      $d_{Rmean}$              & \textit{\OurMethod{}/FT}           & 1.162 & 1.150 & 1.062 & 0.473 & 0.965 & 0.539 & 1.182 & 1.236 & 1.155 \\ 
                 &      $\mathit{\%_{closer}}$              & \textit{\OurMethod{} \textless FT} & 29.4\% & 22.6\% & 44.8\% & 82.7\% & 68.0\% & 84.8\% & 65.8\% & 36.2\% & 66.9\% \\
\midrule

                 &      	$d_{mean}$              & \textit{FT}                & \textbf{0.119}  & 0.028  & 0.126\rlap\NoExample  & \textbf{0.050}  & 0.027  & 0.131\rlap\NoExample  & \textbf{0.121}  & 0.030  & 0.142\rlap\NoExample  \\
   \textit{Shopping}              &              & \textit{\OurMethod{}}              & 0.142\rlap{\dubbelop}  & \textbf{0.025}\rlap{\dubbelneer}  & \textbf{0.028}\rlap{\dubbelneer}  & 0.055\rlap{\dubbelop}  & \textbf{0.013}\rlap{\dubbelneer}  & \textbf{0.006}\rlap{\dubbelneer}  & 0.128\rlap{\notsig}  & \textbf{0.026}\rlap{\dubbelneer}  & \textbf{0.046}\rlap{\dubbelneer}  \\ 
\cmidrule{2-12}
                 &      $d_{Rmean}$              & \textit{\OurMethod{}/FT}           & 1.051  & 1.053  & 0.218  & 0.795  & 0.482  & 0.074  & 0.944  & 0.796  & 0.312  \\ 
                 &    $\mathit{\%_{closer}}$                & \textit{\OurMethod{} \textless FT} & 40.2\% & 36.1\% & 99.6\% & 44.4\% & 86.1\% & 99.5\% & 55.8\% & 81.9\% & 97.1\% \\

 \bottomrule                
\end{tabular*}
\caption{Evaluation results for Experiment 1 comparing \OurMethod{} and FT counterfactual examples. Significant improvements and losses over the baseline (FT) are denoted by \dubbelneer\ and \dubbelop, respectively ($p < 0.05$, two-tailed t-test,); 
\notsig{} denotes no significant difference; \NoExample{} denotes settings where the baseline cannot find a counterfactual example for every instance.}
\label{table:distances}
\end{table*}

\begin{table*}[h!]
\centering
\setlength{\tabcolsep}{3pt}
\begin{tabular}{ll@{}rrrrrr}
\toprule
                &                            & \multicolumn{1}{c}{\textbf{Wine}} & \multicolumn{1}{c}{\textbf{HELOC}} & \multicolumn{2}{c}{\textbf{COMPAS}}                               & \multicolumn{2}{c}{\textbf{Shopping}}                             \\
                \cmidrule(r){3-3}\cmidrule(r){4-4}\cmidrule(r){5-6}\cmidrule{7-8}
\textbf{Metric} & \textbf{Method}            & \multicolumn{1}{c}{\textbf{DT}}   & \multicolumn{1}{c}{\textbf{DT}}    & \multicolumn{1}{c}{\textbf{DT}} & \multicolumn{1}{c}{\textbf{AB}} & \multicolumn{1}{c}{\textbf{DT}} & \multicolumn{1}{c}{\textbf{AB}} \\ \midrule

$d_{mean}$           & \textit{DACE}              & 1.325                             & {1.427}                              & 0.814                           & 1.570                           & 0.050                           & 3.230                           \\
                & \textit{FOCUS}             & \textbf{0.542}\rlap{\dubbelneer}                             & \textbf{0.810}\rlap{\dubbelneer}                              & \textbf{0.776}\rlap{\notsig}                           & \textbf{0.636}\rlap{\dubbelneer}                           & \textbf{0.023}\rlap{\dubbelneer}                           & \textbf{0.303}\rlap{\dubbelneer}                           \\ \midrule
$d_{Rmean}$          & \textit{FOCUS /}           & \multirow{2}{*}{0.420}            & \multirow{2}{*}{0.622}             & \multirow{2}{*}{1.18}          & \multirow{2}{*}{0.372}          & \multirow{2}{*}{0.449}          & \multirow{2}{*}{0.380}          \\
                & \textit{DACE}              &                                   &                                    &                                 &                                 &                                 &                                 \\ \midrule
$\mathit{\%_{closer}}$          & \textit{FOCUS \textless{}} & \multirow{2}{*}{100\%}            & \multirow{2}{*}{94.5\%}             & \multirow{2}{*}{29.9\%}          & \multirow{2}{*}{96.1\%}          & \multirow{2}{*}{99.4\%}          & \multirow{2}{*}{90.8\%}          \\
                & \textit{DACE}              &                                   &                                    &                                 &                                 &                                 &                                 \\ 
\midrule
 \textit{\# CFs}        & \textit{DACE}              & 241                               & 1,342                               & 842                             & 700                             & 362                             & 448                             \\
  \textit{found}              & \textit{FOCUS}             & 1,470                              & 3,138                               & 1,852                            & 1,852                            & 3,699                            & 3,699                            \\ \midrule 
                
   \textit{\# obs      in}	& \textit{dataset}	& 1,470                              & 3,138                               & 1,852                            & 1,852                            & 3,699                            & 3,699                            \\ \bottomrule
\end{tabular}
\caption{Evaluation results for Experiment 2 comparing \OurMethod{} and DACE counterfactual examples in terms of Mahalanobis distance. Significant improvements over the baseline are denoted by \dubbelneer\ ($p < 0.05$, two-tailed t-test,). 
\notsig{} denotes no significant difference.}
\label{table:experiment2}
\end{table*}

\section{Experiment 1: FOCUS vs. FT}
\label{section:experiment1}

We compare \OurMethod{} to the Feature Tweaking (FT) method by \citet{tolomei_interpretable_2017} in terms of the evaluation metrics in Section~\ref{section:evaluation}. 
We consider \numprint{36} experimental settings (\numprint{4} datasets $\times$ \numprint{3} tree-based models $\times$ \numprint{3} distance functions) when comparing \OurMethod{} to FT. 
The results are listed in Table 1. 

In terms of $d_\mathit{mean}$, \OurMethod{} outperforms \ac{FT} in \numprint{20} settings while \ac{FT} outperforms \OurMethod{} in \numprint{8} settings. The difference in $d_\mathit{mean}$ is not significant in the remaining \numprint{8} settings. 
In general, \OurMethod{} outperforms \ac{FT} in settings using Euclidean and Cosine distance because in each iteration, \OurMethod{} perturbs many of the features by a small amount. 
Since \ac{FT} perturbs only the features associated with an individual leaf, we expected that it would perform better for Manhattan distance but our results show that this is not the case -- there is no clear winner between FT and \OurMethod{} for Manhattan distance. 
We also see that \OurMethod{} usually outperforms \ac{FT} in settings using Random Forests (RF) and Adaptive Boosting (AB), while the opposite is true for Decision Trees (DT). 

Overall, we find that \OurMethod{} is effective and efficient for finding counterfactual explanations for tree-based models.
Unlike the \ac{FT} baseline, \OurMethod{} finds valid counterfactual explanations for \emph{every} instance across all settings. 
In the majority of tested settings, \OurMethod{}'s explanations are substantial improvements in terms of distance to the original inputs, across all three metrics.

\section{Experiment 2: FOCUS vs. DACE}
\label{section:experiment2}
The flexibility of FOCUS allows us to plug in our choice of differentiable distance function. To compare against DACE~\citep{kanamori_dace_2020}, we use the Mahalanobis distance for both 
\begin{inparaenum}[(i)]
\item generation of FOCUS explanations, and
\item evaluation in comparison to DACE, since this is the distance function used in the DACE loss function (see Equation~\ref{eq:daceloss} in Section~\ref{section:baselinedace}). 
\end{inparaenum}

Table 2 shows the results for the \numprint{6} settings we could run DACE on. 
We were only able to run DACE on \numprint{6} out of our \numprint{12} models because the problem size is too large (i.e., DACE has too many model parameters) for the remaining \numprint{6} models when using the free Python API of CPLEX (the optimizer used in DACE). 
Therefore, when comparing against DACE, we have \numprint{6} experimental settings (\numprint{6} models $\times$ \numprint{1} distance function).

We found that DACE can only generate counterfactual examples for a small subset of the test set, regardless of the $\lambda$-value, as opposed to \OurMethod{}, which can generate counterfactual examples for the entire test set in all cases. 
To compute $d_{mean}$, $d_{Rmean}$, and $\mathit{\%_{closer}}$, we compare \OurMethod{} and DACE only on the instances for which DACE was able to generate a counterfactual example. 
We find that \OurMethod{} significantly outperforms DACE in \numprint{5} out of \numprint{6} settings in terms of all three evaluation metrics, indicating that FOCUS explanations are indeed more minimal than those produced by DACE. 
FOCUS is also more reliable since 
\begin{inparaenum}[(i)]
\item it is not restricted by model size, and
\item it can generate counterfactual examples for all instances in the test set. 
\end{inparaenum}

\section{Discussion and Analysis}
\label{section:case-study}

\begin{figure}[t]
\begin{center}
\includegraphics[scale=0.39]{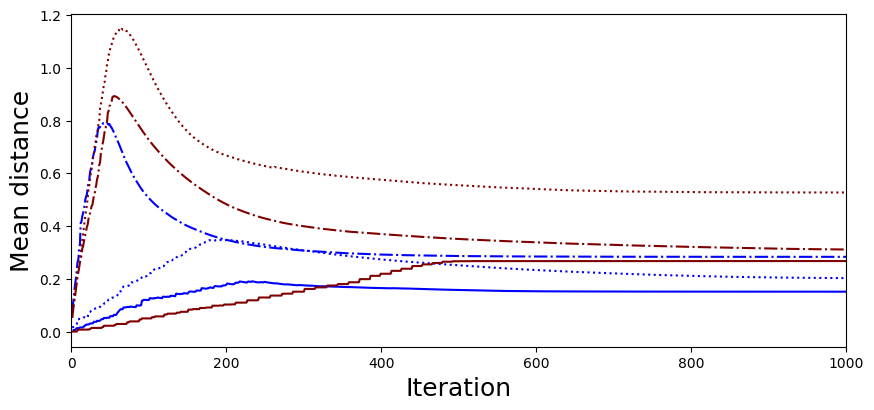} 
\includegraphics[scale=0.39]{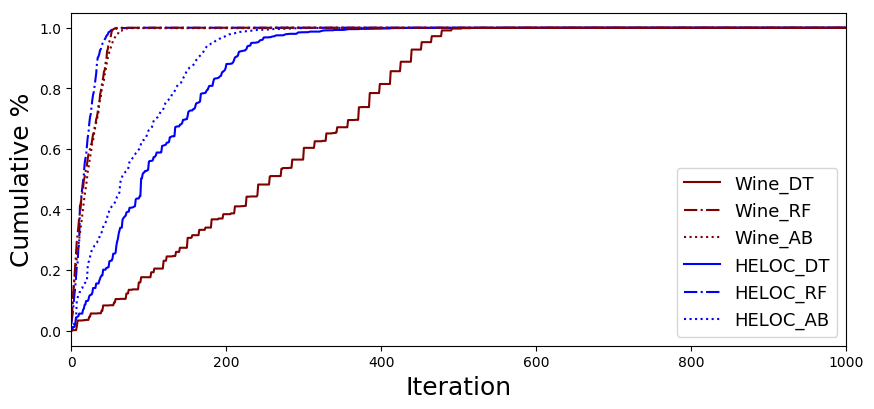} 
\end{center}
\caption{Mean distance (top) and cumulative \% (bottom) of counterfactual examples in each iteration of \OurMethod{} for Manhattan explanations.}
\label{fig:distances}
\end{figure}

Figure~\ref{fig:distances} shows the mean Manhattan distance of the perturbed examples in each iteration of FOCUS, along with the proportion of perturbations resulting in valid counterfactual examples found for two datasets (we omit the others due to space considerations). These trends are indicative of all settings: the mean distance increases until a counterfactual example has been found for every $x$, after which the mean distance starts to decrease. This seems to be a result of the hinge-loss in FOCUS, which first prioritizes finding a valid counterfactual example (see Equation 1), then decreasing the distance between $x$ and $\bar{x}$. 

\begin{figure*}[h!]
\centering
\includegraphics[scale=0.33]{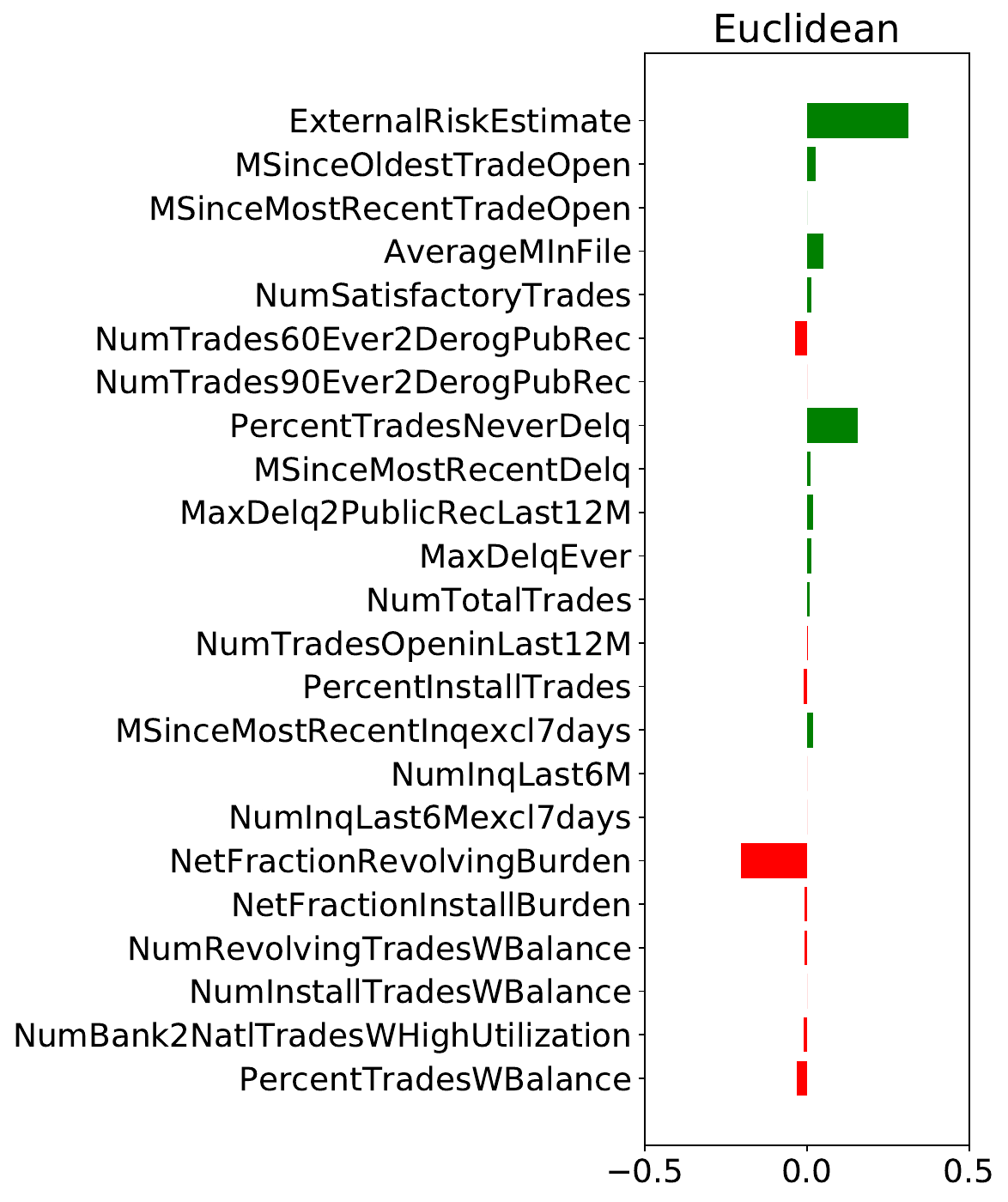}
\includegraphics[scale=0.33]{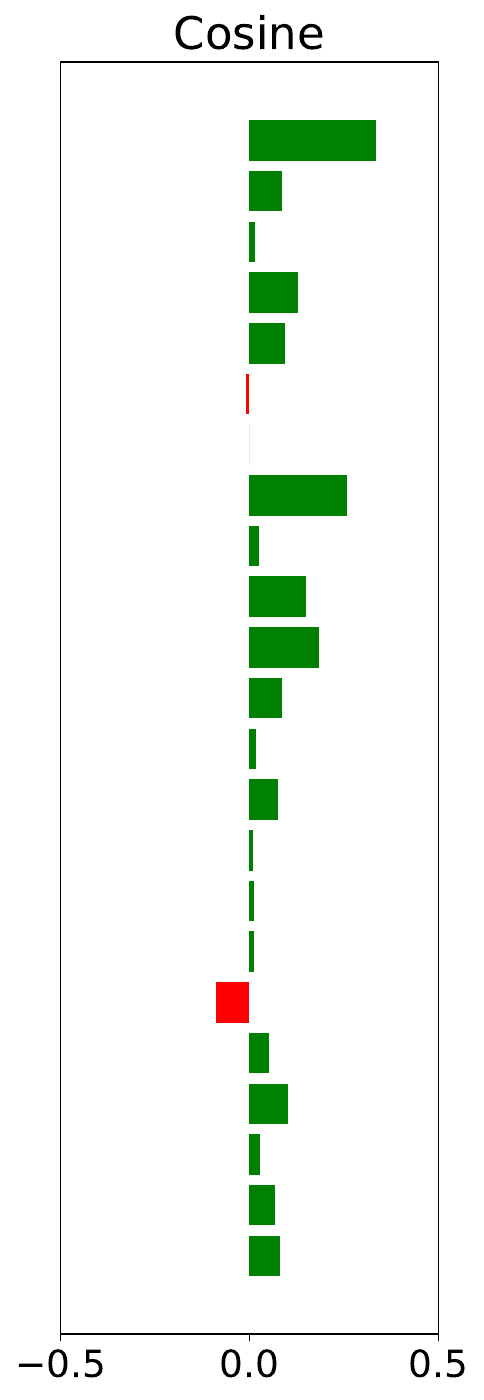} 
\includegraphics[scale=0.33]{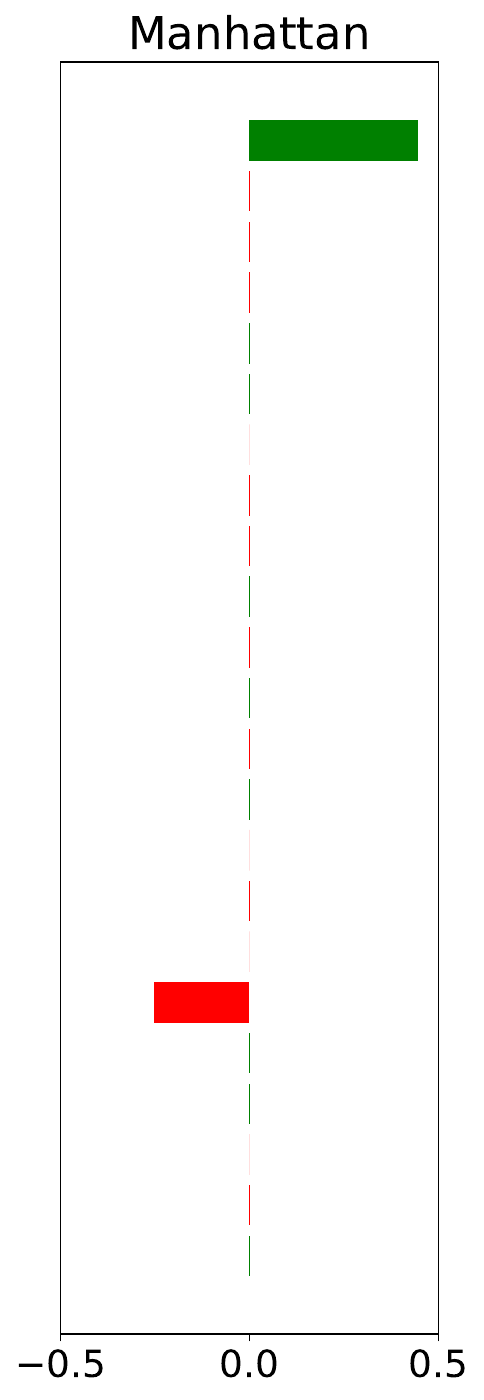} 
\includegraphics[scale=0.33]{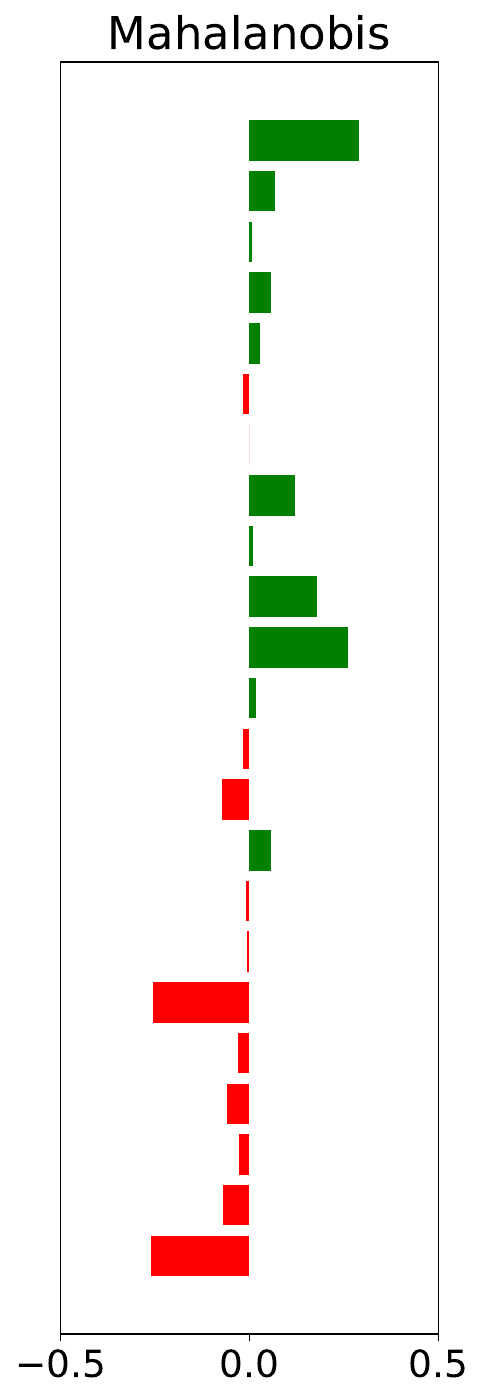} 
\caption{\OurMethod{} explanations for the same model and same $x$ based on different distance functions. 
Green and red indicate increases and decreases in feature values, respectively. 
Perturbation values are based on normalized feature values. 
Left: Euclidean explanation perturbs several features, but only slightly. 
Middle Left: Cosine explanation perturbs almost all of the features. 
Middle Right: Manhattan explanation perturbs two features substantially.
Right: Mahalanobis explanation perturbs almost all of the features. 
}
\label{fig:perturb-examples}
\end{figure*}

\subsection{Case Study: Credit Risk}
As a practical example, we investigate what \OurMethod{} explanations look like for individuals in the HELOC dataset. 
Here, the task is to predict whether or not an individual will default on their loan. 
This has consequences for loan approval: individuals who are predicted as defaulting will be denied a loan. 
For these individuals, we want to understand how they can change their profile such that they are approved. 
Given an individual who has been denied a loan from a bank, a counterfactual explanation could be:
\begin{quote}
\textit{Your loan application has been denied. In order to have your loan application approved, you need to 
\begin{inparaenum}[(i)]
	\item increase your ExternalRiskEstimate score by 62, and
	\item decrease your NetFractionRevolvingBurden by 58.
\end{inparaenum}}
\end{quote}

\noindent%
Figure~\ref{fig:perturb-examples} shows four counterfactual explanations generated using different distance functions for the same individual and same model. 
We see that the Manhattan explanation only requires a few changes to the individual's profile, but the changes are large.
In contrast, the individual changes in the Euclidean explanation are smaller but there are more of them. 
In settings where there are significant dependencies between features, the Cosine explanations may be preferred since they are based on perturbations that try to preserve the relationship between features. 
For instance, in the \textit{Wine Quality} dataset, it would be difficult to change the amount of citric acid without affecting the pH level. 
The Mahalanobis explanations would be useful when it is important to take into account not only correlations between features, but also the training data distribution. 
This flexibility allows users to choose what kind of explanation is best suited for their problem. 

Different distance functions can result in different \emph{magnitudes} of feature perturbations as well as different \emph{directions}. For example, the Cosine explanation suggests increasing \textit{PercentTradesWBalance}, while the Mahalanobis explanations suggests decreasing it. 
This is because the loss space of the underlying RF model is highly non-convex, and therefore there is more than one way to obtain an alternative prediction. When using complex models such as tree ensembles, there are no monotonicity guarantees. In this case, both options result in valid counterfactual examples. 

We examine the Manhattan explanation in more detail. 
We see that \OurMethod{} suggests two main changes: 
\begin{inparaenum}[(i)]
	\item increasing the \textit{ExternalRiskEstimate}, and 
	\item decreasing the \textit{NetFractionRevolvingBurden}. 
\end{inparaenum}
We obtain the definitions and expected trends from the data dictionary \cite{fico-data-dict} created by the authors of the dataset. 
The \emph{ExternalRiskEstimate} is a ``consolidated version of risk markers'' (i.e., a credit score). 
A higher score is better: as one's \emph{ExternalRiskEstimate} increases, the probability of default decreases. 
The \textit{NetFractionRevolvingBurden} is the ``revolving balance divided by the credit limit'' (i.e., utilization). 
A lower value is better: as one's \emph{NetFractionRevolvingBurden} increases, the probability of default increases. 
We find that the changes suggested by \OurMethod{} are fairly consistent with the expected trends in the data dictionary \cite{fico-data-dict}, as opposed to suggesting nonsensical changes such as increasing one's utilization to decrease the probability of default.

Decreasing one's utilization is heavily dependent on the specific situation: an individual who only supports themselves might have more control over their spending in comparison to someone who has multiple dependents. 
An individual can decrease their utilization in two ways: 
\begin{inparaenum}[(i)]
	\item decreasing their spending, or
	\item increasing their credit limit (or a combination of the two).
\end{inparaenum}
We can postulate that (i) is more ``actionable'' than (ii), since (ii) is usually a decision made by a financial institution. 
However, the degree to which an individual can actually change their spending habits is completely dependent on their specific situation: an individual who only supports themselves might have more control over their spending than someone who has multiple dependents. 
In either case, we argue that deciding what is (not) actionable is not a decision for the developer to make, but for the individual who is affected by the decision. 
Counterfactual examples should be used as part of a human-in-the-loop system and not as a final solution. 
The individual should know that utilization is an important component of the model, even if it is not necessarily ``actionable'' for them. 
We also note that it is unclear how exactly an individual would change their credit score without further insight into how the score was calculated (i.e., how the risk markers were consolidated).
It should be noted that this is not a shortcoming of \OurMethod{}, but rather of using features that are uninterpretable on their own, such as credit scores.
Although \OurMethod{} explanations cannot tell a user precisely how to increase their credit score, it is still important for the individual to know that their credit score is an important factor in determining their probability of getting a loan, as this empowers them to ask questions about how the score was calculated (i.e., how the risk markers were consolidated).

\section{Conclusion}
\label{section:conclusion}
We propose an explanation method for tree-based classifiers, \OurMethod{}, which casts the problem of finding counterfactual examples as a gradient-based optimization task and provides a differentiable approximation of tree-based models to be used in the optimization framework. 
Given an input instance $x$, \OurMethod{} generates an optimal counterfactual example based on the minimal perturbation to the input instance $x$ which results in an alternative prediction from a model $f$. 
Unlike previous methods that assume the underlying classification model is differentiable, we propose a solution for when $f$ is a non-differentiable, tree-based model that provides a differentiable approximation of $f$ that can be used to find counterfactual examples using gradient-based optimization techniques.  
In the majority of experiments, examples generated by \OurMethod{} are significantly closer to the original instances in terms of three different evaluation metrics compared to those generated by the baselines. 
\OurMethod{} is able to generate valid counterfactual examples for all instances across all datasets, and the resulting explanations are flexible depending on the distance function. 
We plan to conduct a user study to test how varying the distance functions impacts user preferences for explanations.

\section*{Reproducibility}
To facilitate the reproducibility of this work, our code is available at \url{https://github.com/a-lucic/focus}.

\section*{Acknowledgements}
This research was supported by Ahold Delhaize and the Netherlands Organisation for Scientific Research under project nr.\ 652.\-001.\-003., and by the Hybrid Intelligence Center,  a 10-year program funded by the Dutch Ministry of Education, Culture and Science through the Netherlands Organisation for Scientific Research,  \url{https://hybrid-intelligence-centre.nl}, and by the Google Research Scholar Program.

All content represents the opinion of the authors, which is not necessarily shared or endorsed by their respective employers and/or sponsors.

\bibliographystyle{aaai22}
\bibliography{thesis}

\end{document}


\maketitle

\appendix

\section{Models DACE Can't Run On}
\label{section:dace-failures}

We were unable to run DACE on the following settings because the problem size is too large when using the free Python API of CPLEX:
\begin{itemize}[leftmargin=*]
	\item Wine Quality AB (100 trees, max depth 4)
	\item Wine Quality RF (500 trees, max depth 4)
	\item HELOC RF (500 trees, max depth 4)
	\item HELOC AB (100 trees, max depth 8)
	\item COMPAS RF (500 trees, max depth 4)
	\item Shopping RF (500 trees, max depth 8).
\end{itemize}
We note that these are not unreasonable model sizes, and that unlike DACE, \OurMethod{} can be applied to all \numprint{12} models (see Table 1 in Section 5).

\section{Evaluation Metrics}
\label{section:metrics}
\vspace{5mm}
Here we show exact calculations for mean distance ($d_\mathit{mean}$) and mean relative distance ($d_\mathit{Rmean}$). 
Let $X$ be the set of $N$ original instances and $\bar{X}$ be the corresponding set of $N$ generated counterfactual examples.
The mean distance is defined as:
\begin{equation}
\label{eq:mean-dist}
d_\mathit{mean}(X, \bar{X}) = \frac{1}{N}\sum_{n=1}^{N}d(x^{(n)}, \bar{x}^{(n)}).
\end{equation}

Let $\bar{X}$ be the set of counterfactual examples produced by \OurMethod{} and let $\bar{X}'$ be the set of counterfactual examples produced by a baseline. 
Then the mean relative distance is defined as:
\begin{equation}
\label{eq:mean-rel-dist}
d_\mathit{Rmean}(\bar{X}, \bar{X}') = \frac{1}{N}\sum_{n=1}^{N} \frac{d(x^{(n)}, \bar{x}^{(n)})}{d(x^{(n)}, {\bar{x}}^{'(n)})}.
\end{equation}

\section{Fidelity of Approximations}
Fidelity is commonly used to evaluate XAI methods that are based on approximations of a given model: it is a measure of the agreement between the original model $\mathcal{M}$ and its approximation $\widetilde{\mathcal{M}}$: 
\begin{align}
\begin{split}
\textit{fid}(\widetilde{\mathcal{M}}, X) \!=\! 
\frac{1}{N} \sum_{n=1}^N
\mathbbm{1}\left[
y_{x^{(n)}}
\!=\! \arg\max_{y'} \widetilde{\mathcal{M}}(y' | x^{(n)})
\right],
\end{split}
\end{align}
where $y_x = \arg\max_{y} \mathcal{M}(y \mid x)$. 
In our case, the purpose of the approximations is not to replace the original models, but rather to construct differentiable versions of the models so we can generate counterfactual examples through gradient-based optimization. 
Examining the fidelity is meant as a \emph{sanity check} -- to ensure our approximations are reasonably representative of the original model.

Table~\ref{table:fidelity} shows the fidelity of the model approximations used in our experiments: a value of \numprint{1} indicates perfect alignment between $\mathcal{M}$ and $\widetilde{\mathcal{M}}$. 
We see that the alignment is at least \numprint{0.7}, which indicates that \OurMethod{} approximations are indeed reasonable representations of the original model -- both in terms of their inner workings (i.e., same tree structure, same features, same splitting thresholds but ``softer'' versions) as well as their predictions.

\begin{table}
\centering
\caption{Fidelity of FOCUS approximations used in Experiments \numprint{1} and \numprint{2}: --- denotes models we were unable to run DACE on.
The parameters $\sigma$ and $\tau$ are chosen per setting based on which values produce the best counterfactual examples.
As a result, different approximations of the same model are used when optimizing for different distance functions, but in some cases (e.g., \textit{Wine Quality} DT), the approximations are the same regardless of the distance function.}

\begin{tabular}{lcrrrr}
\toprule
\textbf{Dataset}              & \textbf{Model} & \multicolumn{1}{c}{\textbf{Euclid.}} & \multicolumn{1}{c}{\textbf{Cosine}} & \multicolumn{1}{c}{\textbf{Man.}} & \multicolumn{1}{c}{\textbf{Mah.}} \\
\midrule
\textit{Wine} & DT             & 0.836     & 0.836  & 0.836    &	0.900\\

    \textit{Quality}                          & RF             		& 0.940     & 0.940  & 0.940  &  --- \\
                              & AB       			& 0.926     & 0.926  & 0.926   &  --- \\
\midrule
\multirow{3}{*}{\textit{HELOC}}        & DT             & 0.836     & 0.836  & 0.836  &   0.930 \\
                              & RF             		& 0.954     & 0.887  & 0.887   &  --- \\
                              & AB       			& 0.936     & 0.744  & 0.905    & --- \\
\midrule
\multirow{3}{*}{\textit{COMPAS}} & DT    	& 0.844     & 0.894  & 0.807   &  0.807 \\
						& RF 	& 0.742     & 0.809  & 0.700   &  --- \\
						& AB		& 0.922     & 0.922  & 0.814   &  0.814 \\
						
\midrule
\multirow{3}{*}{\textit{Shopping}} & DT    	& 0.902     & 0.906  & 0.902  &   0.902 \\
						& RF 	& 0.810     & 0.780  & 0.871   &  --- \\
						& AB		& 0.919     & 0.919  & 0.919	& 0.919	\\
\bottomrule     
\end{tabular}
\label{table:fidelity}
\end{table}

\pagebreak

\section{Reproducibility Checklist}
Here we provide detailed answers to the reproducibility checklist. 

\subsection{This paper explains how the results substantiate the claims}
Yes, see Sections 5 and 6. 

\subsection{This paper explicitly identifies limitations or technical assumptions}
Yes, see Section 7. 

\subsection{This paper includes a conceptual outline and/or pseudocode description of AI methods introduced.}
Yes, see Section 3.

\subsection{Does this paper make theoretical contributions?}
No. 

\subsection{Does this paper rely on one or more data sets?}
Yes. All datasets are drawn from existing literature and are publicly available. There are no novel datasets.

\subsection{Does this paper include computational experiments?}
\begin{itemize}
	\item Yes. All source code is included in the appendix and is available in an anonymous online repository: https://anonymous.4open.science/r/48c9c410-61ed-4e19-aecf-79c588122dc0/. 
	\item The method is deterministic, so there is no randomness involved and there is only one run per result. 
	\item The experiments were conducted in Python 3.7, using Tensorflow 1.14, on a 48-core CPU with 256GB of RAM. The environment file with all required software libraries is available in the online repository. 
	\item Information about hyperparameters can be found in Appendix~\ref{section:hyperparameters}. 
\end{itemize}

\section{FOCUS Hyperparameters}
\label{section:hyperparameters}
Here we detail the FOCUS hyperparameters across the \numprint{42} settings in Experiments 1 and 2: $\sigma$ indicates the steepness of the sigmoid function in Equation 14; $\tau$ is the temperature of the softmax in Equation 16; $\beta$ is the trade-off parameter in Equation 7; $\alpha$ is the learning rate of Adam. 

\begin{table*}
\centering
\caption{FOCUS hyperparameters used in Experiment 1 comparing \OurMethod{} with FT and RP using Euclidean distance.}
\label{table:hyperparameters-euc}
\begin{tabular}{lcccrrrr}
\toprule
\textbf{Dataset}                   & \textbf{Model}       & \textbf{Num Trees}   & \textbf{Max Depth}   & $\sigma$ & $\tau$ & $\beta$ & $\alpha$ \\
\textit{Wine}                      & DT                   & 1                                  & 2                                  & 1  & 10 & 0.05 & 0.001    	 \\
\textit{Quality}                   & RF                   & 500                                & 4                                  & 10 & 2  & 0.05 & 0.005      	 \\
\textit{}                          & AB                   & 100                                & 4                                  & 5  & 1  & 0.05 & 0.005      	 \\ \midrule
\multirow{3}{*}{\textit{HELOC}}    & DT                   & 1                                  & 4                                  & 2  & 10 & 0.05 & 0.001	    \\
                                   & RF                   & 500                                & 4                                  & 10 & 5  & 0.05 & 0.005      	\\
                                   & AB                   & 100                                & 8                                  & 10 & 1  & 0.05 & 0.001  	   \\ \midrule
\multirow{3}{*}{\textit{COMPAS}}   & DT                   & 1                                  & 4                                  & 6  & 10 & 0.05 & 0.005       \\
                                   & RF                   & 500                                & 4                                  & 7  & 3  & 0.01 & 0.001      \\
                                   & AB                   & 100                                & 2                                  & 10 & 1  & 0.01 & 0.005	\\ \midrule
\multirow{3}{*}{\textit{Shopping}} & DT                   & 1                                  & 4                                  & 2  & 10 & 0.05 & 0.005     \\
                                   & RF                   & 500                                & 8                                  & 5  & 5  & 0.05 & 0.005      \\
                                   & AB                   & 100                                & 2                                 & 10 & 1  & 0.05 & 0.001      \\ \bottomrule
\end{tabular}
\end{table*}

\begin{table*}
\centering
\caption{FOCUS hyperparameters used in Experiment 1 comparing \OurMethod{} with FT and RP using Cosine distance.}
\label{table:hyperparameters-cos}
\begin{tabular}{lcccrrrr}
\toprule
\textbf{Dataset}                   & \textbf{Model}       & \textbf{Num Trees}   & \textbf{Max Depth}   & $\sigma$ & $\tau$ & $\beta$ & $\alpha$ \\ \midrule
\textit{Wine}                      & DT                   & 1                                  & 2                                  & 1  & 10 & 0.05 & 0.005      	 \\
\textit{Quality}                   & RF                   & 500                                & 4                                  & 10 & 1  & 0.05 & 0.005      	 \\
\textit{}                          & AB                   & 100                                & 4                                  & 1  & 1  & 0.01 & 0.005      	 \\ \midrule
\multirow{3}{*}{\textit{HELOC}}    & DT                   & 1                                  & 4                                  & 2  & 10 & 0.05 & 0.005  	    \\
                                   & RF                   & 500                                & 4                                  & 5  & 5  & 0.05 & 0.005       	\\
                                   & AB                   & 100                                & 8                                  & 1  & 1  & 0.05 & 0.005   	   \\ \midrule
\multirow{3}{*}{\textit{COMPAS}}   & DT                   & 1                                  & 4                                  & 10 & 10 & 0.05 & 0.005       \\
                                   & RF                   & 500                                & 4                                  & 10 & 6  & 0.01 & 0.005       \\
                                   & AB                   & 100                                & 2                                  & 10 & 1  & 0.05 & 0.005       	\\ \midrule
\multirow{3}{*}{\textit{Shopping}} & DT                   & 1                                  & 4                                  & 10 & 10 & 0.05 & 0.001      \\
                                   & RF                   & 500                                & 8                                  & 1  & 1  & 0.01 & 0.001       \\
                                   & AB                   & 100                                & 2                                  & 10 & 5  & 0.05 & 0.001      \\ \bottomrule
\end{tabular}
\end{table*}


\begin{table*}
\centering
\caption{FOCUS hyperparameters used in Experiment 1 comparing \OurMethod{} with FT and RP using Manhattan distance.}
\label{table:hyperparameters-manhat}
\begin{tabular}{lcccrrrr}
\toprule
\textbf{Dataset}                   & \textbf{Model}       & \textbf{Num Trees}   & \textbf{Max Depth}   & $\sigma$ & $\tau$ & $\beta$ & $\alpha$ \\ \midrule
\textit{Wine}                      & DT                   & 1                                  & 2                                  & 1  & 10 & 0.05 & 0.001     	 \\
\textit{Quality}                   & RF                   & 500                                & 4                                  & 10 & 10 & 0.01 & 0.005    	 \\
\textit{}                          & AB                   & 100                                & 4                                  & 6  & 1  & 0.01 & 0.005     	 \\ \midrule
\multirow{3}{*}{\textit{HELOC}}    & DT                   & 1                                  & 4                                 & 2  & 10 & 0.05 & 0.001  	    \\
                                   & RF                   & 500                                & 4                                 & 5  & 5  & 0.01 & 0.005     	\\
                                   & AB                   & 100                                & 8                                  & 4  & 1  & 0.05 & 0.001  	   \\ \midrule
\multirow{3}{*}{\textit{COMPAS}}   & DT                   & 1                                  & 4                                  & 6  & 10 & 0.01 & 0.005      \\
                                   & RF                   & 500                                & 4                                   & 4  & 1  & 0.05 & 0.001       \\
                                   & AB                   & 100                                & 2                                  & 5  & 10 & 0.05 & 0.005  	\\ \midrule
\multirow{3}{*}{\textit{Shopping}} & DT                   & 1                                  & 4                                  & 2  & 10 & 0.05 & 0.005    \\
                                   & RF                   & 500                                & 8                                  & 10 & 1  & 0.05 & 0.001       \\
                                   & AB                   & 100                                & 2                                  & 10 & 1  & 0.05 & 0.001     \\ \bottomrule
\end{tabular}
\end{table*}


\begin{table*}
\centering
\caption{FOCUS hyperparameters used in Experiment 2 comparing \OurMethod{} with DACE using Mahalanobis distance.}
\label{table:hyperparameters-mahal}
\begin{tabular}{lcccrrrr}
\toprule
\textbf{Dataset}                   & \textbf{Model}       & \textbf{Num Trees}   & \textbf{Max Depth}   & $\sigma$ & $\tau$ & $\beta$ & $\alpha$ \\ \midrule
\textit{Wine Quality}                      & DT                   & 1                                  & 2                                  & 5  & 10 & 0.01  & 0.001       	 \\ \midrule
\textit{HELOC}   & DT                   & 1                                  & 4                                  & 5  & 10 & 0.01  & 0.001   	    \\ \midrule

\multirow{2}{*}{\textit{COMPAS}}   & DT                   & 1                                  & 4                                  & 5  & 10 & 0.01  & 0.005   \\
                                   & AB                   & 100                                & 2                                  & 4  & 2  & 0.005 & 0.001      	\\ \midrule
\multirow{2}{*}{\textit{Shopping}} & DT                   & 1                                  & 4                                 & 4  & 10 & 0.01  & 0.005 \\
                                   & AB                   & 100                                & 2                                  & 10 & 1  & 0.01  & 0.001      \\ \bottomrule
\end{tabular}
\end{table*}